\documentclass[letterpaper, 10 pt, conference]{ieeeconf}  

\usepackage{comment}
\usepackage{amsmath,amssymb}
\usepackage{color}
\usepackage{url}
\usepackage{hyperref}
\usepackage{graphicx}
\usepackage{amsmath,amssymb}
\usepackage{tikz}
\usepackage{multirow}
\usepackage{cite}
\usepackage{floatrow}
\usepackage{comment}
\usepackage{amsmath,amssymb} 
\usepackage{color}
\usepackage{floatrow}
\usepackage{array}
\usepackage{float}
\usepackage{mathtools}
\usepackage{booktabs}
\floatstyle{plaintop}
\usepackage[export]{adjustbox}
\usepackage{mathtools}

\newcommand{\fun}[3]{\ensuremath{#1\colon #2\mapsto #3}}

\usepackage{amsmath,amsfonts,bm}

\def\eqref#1{equation~\ref{#1}}

\def\1{\bm{1}}

\DeclareMathAlphabet{\mathsfit}{\encodingdefault}{\sfdefault}{m}{sl}
\SetMathAlphabet{\mathsfit}{bold}{\encodingdefault}{\sfdefault}{bx}{n}

\newcommand{\R}{\mathbb{R}}

\restylefloat{table}
\newcolumntype{M}{>{\centering\arraybackslash}m{\dimexpr.22\linewidth-2\tabcolsep}}
\newcolumntype{S}{>{\centering\arraybackslash}m{\dimexpr.05\linewidth-5\tabcolsep}}

\newcommand{\SO}{\ensuremath{\mathbf{SO}}}
\newenvironment{tightitemize} 
{\vspace{-\topsep}\begin{itemize}\itemsep1pt \parskip0pt \parsep0pt}
{\end{itemize}\vspace{-\topsep}}
\DeclareMathOperator*{\argmax}{arg\,max}

\IEEEoverridecommandlockouts                              

\title{\LARGE \bf
Stable Yaw Estimation of Boats from the Viewpoint of UAVs and USVs 
}

\author{Benjamin Kiefer$^{1}$, Timon Höfer$^{1}$ and Andreas Zell$^{1}$
\thanks{$^{1}$All authors are with the Faculty of Computer Science,
        University of Tuebingen, Germany.
        {\tt\small prename.surname@uni-tuebingen.de}}%
}

\begin{document}

\maketitle
\thispagestyle{empty}
\pagestyle{empty}

\begin{abstract}

Yaw estimation of boats from the viewpoint of unmanned aerial vehicles (UAVs) and unmanned surface vehicles (USVs) or boats is a crucial task in various applications such as 3D scene rendering, trajectory prediction, and navigation. However, the lack of literature on yaw estimation of objects from the viewpoint of UAVs has motivated us to address this domain. In this paper, we propose a method based on HyperPosePDF for predicting the orientation of boats in the 6D space. For that, we use existing datasets, such as PASCAL3D+ and our own datasets, SeaDronesSee-3D and BOArienT, which we annotated manually. We extend HyperPosePDF to work in video-based scenarios, such that it yields robust orientation predictions across time. Naively applying HyperPosePDF on video data yields single-point predictions, resulting in far-off predictions and often incorrect symmetric orientations due to unseen or visually different data. To alleviate this issue, we propose aggregating the probability distributions of pose predictions, resulting in significantly improved performance, as shown in our experimental evaluation. Our proposed method could significantly benefit downstream tasks in marine robotics.
 
\end{abstract}

\section{Introduction}

Yaw estimation of objects from the viewpoint of unmanned aerial vehicles (UAVs) and unmanned surface vehicles (USVs) or boats is an essential task in various applications such as 3D scene rendering, trajectory prediction, and navigation. Accurate pose estimation is crucial for safe and efficient operations in the marine environment, where the ability to locate and track objects such as boats and ships is essential for collision avoidance, search and rescue, and marine surveillance. Furthermore, it is vital to have robust yaw predictions in augmented reality applications, to better aid a human operator.

AIS (automatic identification system) data only helps for boats that emit these signals. Smaller boats do not send AIS data. Furthermore, radar is expensive and only provides a very coarse position of boats. It requires a correct set-up of the radar and is harder to interpret for non-experts. Computer vision-based orientation prediction on the other hand offers a cheap and direct method. 

Furthermore, there is a lack of literature on heading estimation of objects from the viewpoint of UAVs and USVs. In particular, predicting the orientation of objects far away from the camera is a challenging task due to the inherent uncertainty in the visual data. Methods based on 3D bounding box detection rely on precise box labels and are inherently error-prone for distant objects \cite{arnold2019survey}.

\begin{figure}
    \centering
    \includegraphics[width=1\textwidth,trim={0 0 0 0},clip]{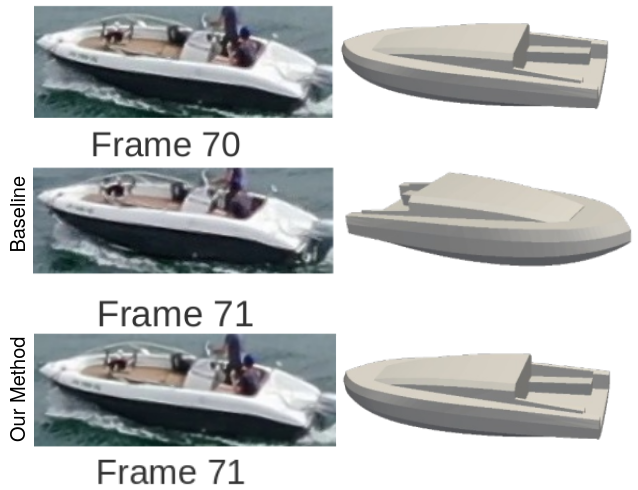}
    \caption{Ignoring the temporal domain results in false, near-symmetric orientation prediction of a boat from frame 70 (top) to frame 71 (middle). Tracking the probability distributions alleviates this problem (bottom). \vspace{-6mm}}
    \label{fig:pose_estimation}
\end{figure}

In this paper, we propose a method based on HyperPosePDF \cite{Hofer_2023_WACV} for predicting the orientation of boats in the 6D space. HyperPosePDF is a recent method that models the uncertainty of predictions and has shown promising results in the field of 6D pose estimation. We train this method on existing datasets, such as PASCAL3D+, and on our own datasets, called SeaDronesSee-3D and BOArienT, which we manually annotate with bounding boxes and pose information for evaluation purposes.

To speed up the bounding box annotation, we develop an annotation tool based on the recently published "Segment Anything" method \cite{kirillov2023segany}. We make this tool together with the data publicly available.

We extend HyperPosePDF to work in video-based scenarios, where the prediction of the orientation of objects across time is essential. Naively applying HyperPosePDF on video data yields single-point predictions, often resulting in far-off predictions and incorrect symmetric orientations due to unseen or visually different data. Therefore, we propose aggregating the probability distributions of pose predictions over time, resulting in significantly improved performance, as shown in our experimental evaluation.

Furthermore, naively predicting the yaw of boats based on analyzing their trajectory in 3D space does not work for standing or slowly moving boats. Moreover, formulating yaw prediction in this way is error-prone due to an ill-posed 2D $\longleftrightarrow$ 3D projection, which is not reliable in heading estimation as we will see in subsequent sections.

Lastly, we demonstrate a full pipeline with detection and tracking of objects and subsequent orientation prediction for a downstream synthetic rendering of a scene. Our proposed method could significantly benefit downstream tasks in marine robotics.

In summary, our contributions are as follows:
\vspace{3mm}

\begin{tightitemize}
    
    \item We pose the novel problem of predicting the heading of boats via purely vision-based methods.
    \item We propose a novel method to aggregate pose predictions by tracking the probability distributions to capture uncertainties due to symmetries and ambiguous appearances.
    \item We create a new dataset BOArienT, a benchmark featuring 30 FPS manually annotated video, featuring precise object detection and pose labels. Furthermore, we annotate parts of SeaDronesSee-MOT with pose data, which we call SeaDronesSee-3D.
    \item We show in multiple experiments on diverse benchmarks the utility of our method. Lastly, we demonstrate the utility of our method on a full pipeline with detection and tracking to synthetically render a scene. 
    \item We make code, data, and adapted labeling tools publicly available on \url{www.macvi.org}.
    
\end{tightitemize}


\section{Related Work}

Pose estimation of common or close industrial objects has been explored in several methods \cite{hofer2021object,labbe2020cosypose,sundermeyer2020augmented}. Analyzing the static images, they split the task into two stages - object detection and subsequent 6D pose estimation of the predicted bounding boxes. However, their focus is on close objects that are dominant in the image plane. On the other hand, we focus on yaw estimation of boats that are distant and occasionally hardly visible. This makes an accurate yaw estimation hard as many plausible predictions exist. Several works explored how to model the uncertainty of pose predictions
\cite{Hofer_2023_WACV,murphy2021implicit}. They output probability distributions over many different poses, effectively capturing the symmetries inherent in the poorly visible objects. While they only experiment with common objects in static scenes, we aim to build on top of their methods to predict stable yaw predictions across time.

The last years have shown a great influx in works in maritime computer vision \cite{varga2021seadronessee,kiefer2023fast,MODSBenchmark2022,kiefer20231st}. Most works focus on the detection or tracking of objects from the viewpoint of UAVs, USVs or boats. There is a great corpus of works working on simulation and trajectory prediction \cite{sullivan2006predictive,browning1991mathematical}. However, these methods only operate on map data as opposed to image/video data.

Likewise, the general UAV-/USV-based research focuses on object detection and tracking, and anomaly detection \cite{du2018unmanned,s23073691,kiefer2023fast,kiefer2021leveraging,messmer2021gaining,kiefer2022leveraging}, but neglects the yaw estimation aspect.


\section{3D Geometry Prerequisites}
\label{sec:3dgeomprereque}

\begin{figure}
    \centering
    \includegraphics[width=0.98\textwidth]{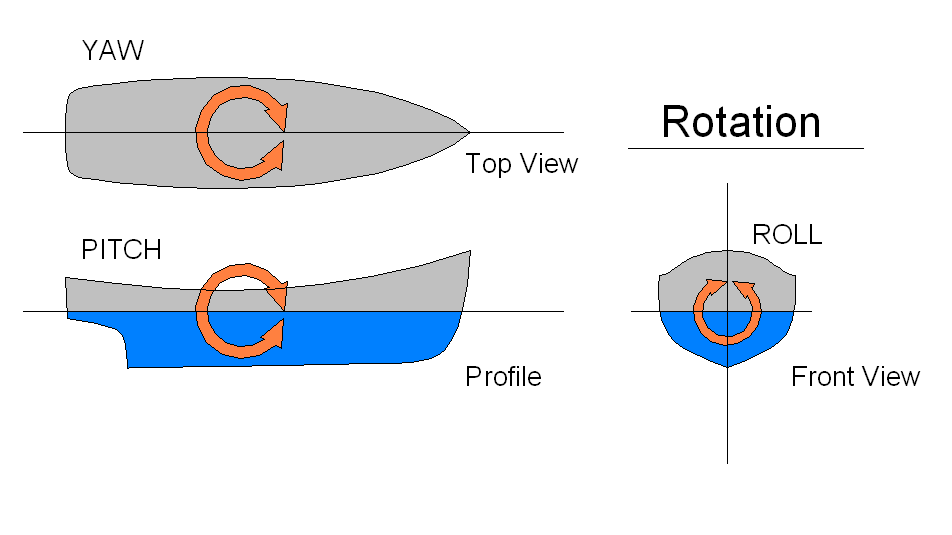}
    \vspace{-5mm}
    \caption{Rotation around the longitudinal, transverse and vertical axes, i.e. roll, pitch, and yaw \cite{yawpitchroll}.}
    \label{fig:yawpitchroll}
    \vspace{-5mm}
\end{figure}

\begin{figure}
    \centering
    \includegraphics[width=0.99\textwidth]{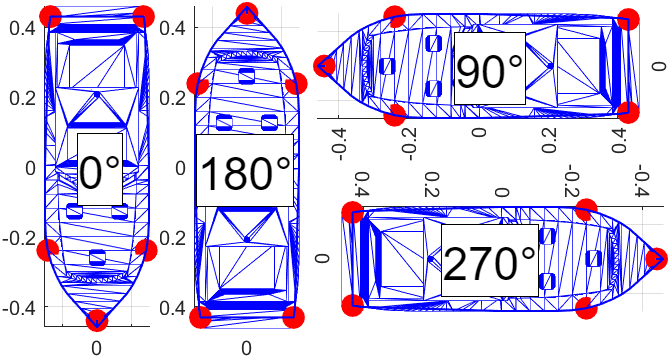}
    \caption{Orientation relative to the camera. At 0$^\circ$, the boat's nose is facing directly us. Note that we did not include the roll and pitch angles.}
    \label{fig:standard_introduction}
\end{figure}

There are three principal axes in any boat, called longitudinal, transverse and vertical axes. Figure \ref{fig:yawpitchroll} shows the rotations around these. These are absolute orientations, i.e. while our method outputs an orientation estimation, it is relative to our camera view. Therefore, we may obtain the absolute orientation using an onboard magnetometer or dual GNSS solutions.

We note that we focus on the case of zero roll and pitch angle, i.e. only the orientation is predicted. 

\begin{figure*}
    \centering
    \includegraphics[trim=0 0 0 0,clip,width=0.49\textwidth]{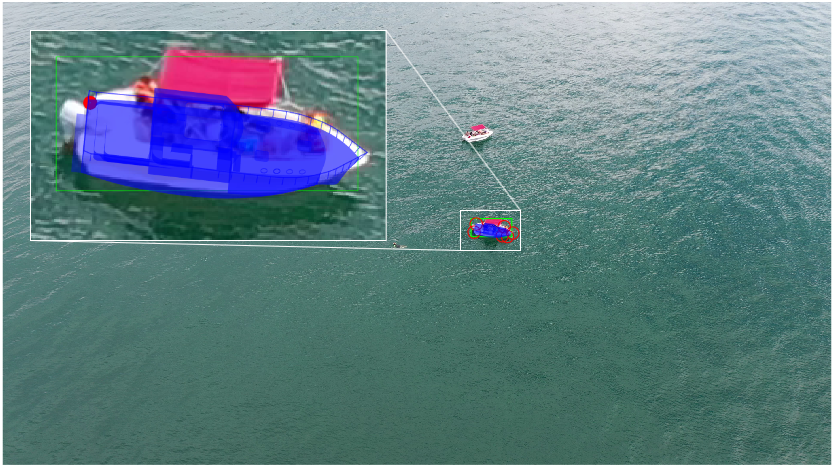}
    \includegraphics[trim=0 0 0 0,clip,width=0.49\textwidth]{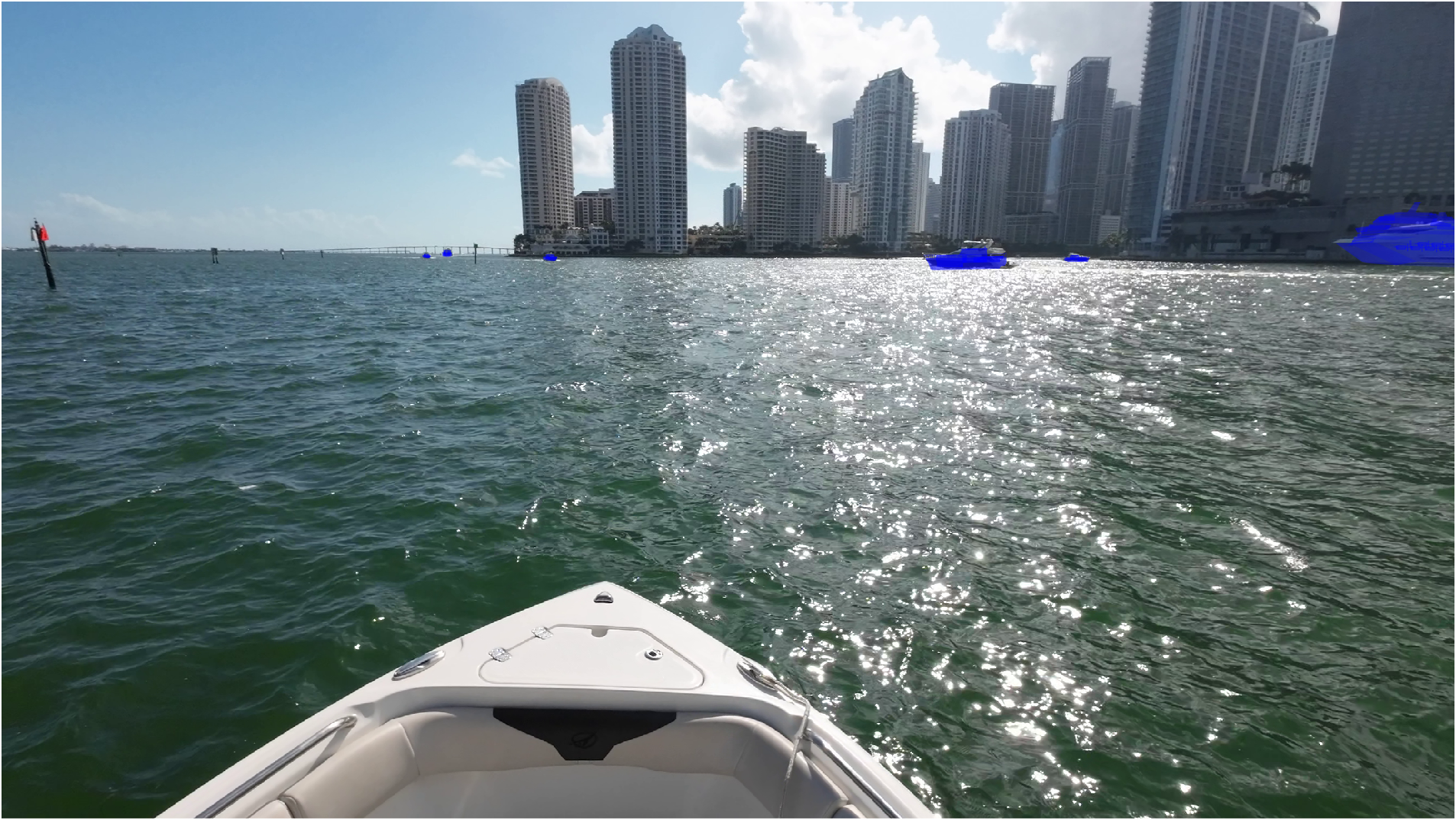}
    \caption{Left: Example orientation of a boat taken from 50m of altitude and looking down with a pitch angle of 40$^\circ$. The highlighted boat has a yaw angle of 280$^\circ$ relative to our viewpoint. Since the UAV's heading is 170$^\circ$ (close to true south), we know that the boat has an absolute heading of 260$^\circ$ (close to true west). Right: Cad overlays on a frame of one the videos we took. Note the very small objects in the left part of the frame.}
    \label{fig:example_orientation}
\end{figure*}

\begin{figure}
    \centering
    \includegraphics[trim=0 0 0 0,clip,width=0.98\textwidth]{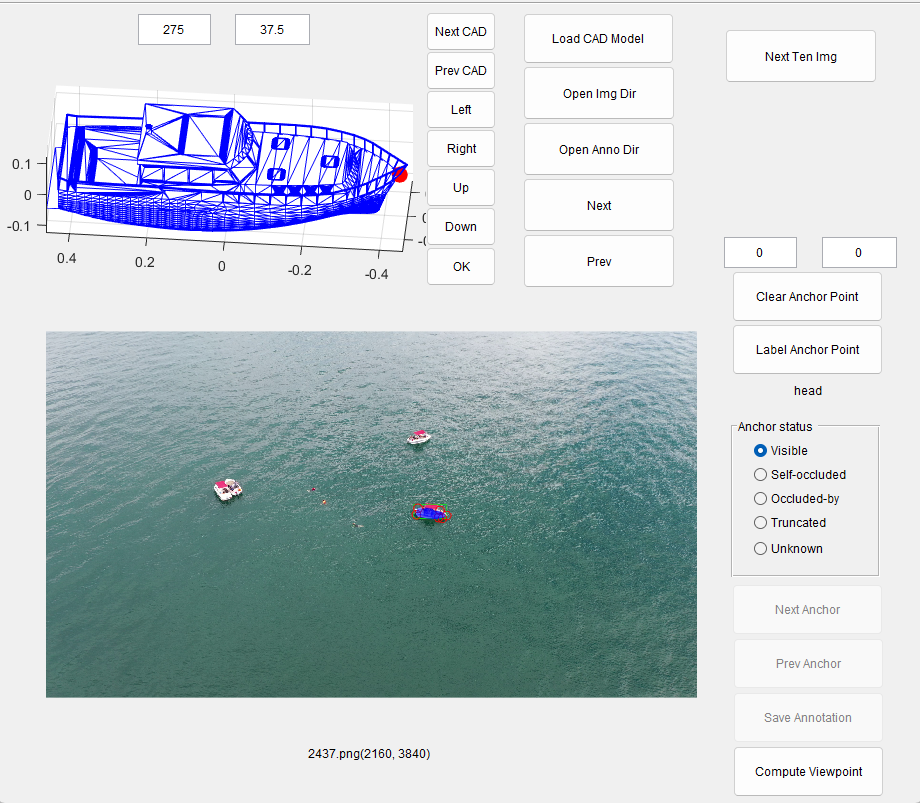}

    \caption{Example view of pose labeling tool. First, we align the view coarsly in steps of 5$^\circ$, then we put the visible anchor points (see Figure \ref{fig:standard_introduction}) in the image plane. These are used to obtain a better pose label. We optimize the orientation to match these anchors (see \cite{murphy2021implicit}).}
    \label{fig:3dlabel_tool}
\end{figure}

For downstream tasks, such as trajectory prediction for collision avoidance but also for rendering synthetic scenes visually smooth and stable, we need to map our predictions to 3D space. For that, we compute 3D object coordinates relative to the UAV, and then use these to obtain actual world coordinates via passive geolocation.

For the relative object coordinates, we consider a mathematical perspective projection camera model since this resembles the common use case for cameras on UAVs and USVs. We assume our camera to look down at a certain angle, which may be a variable gimbal or static camera. A gimbal balances a potential UAV roll angle so that we assume there to be a zero camera roll angle. If there is no gimbal in the USV case, we apply a CV-based roll correction by levelling the horizon line using the IMU roll angle.

Using the relative coordinates of an object ($x$- and $y$- ground distances to UAV), we compute its GPS coordinates based on the UAV's GPS coordinates as follows. Given the camera heading angle $\theta$, we compute the rotation matrix and rotate the relative coordinates of an object to obtain

\begin{equation}
    \begin{bmatrix}
        x_r \\ y_r \\ 1
    \end{bmatrix} = \begin{bmatrix}
\cos (\theta) & -\sin (\theta) & 0\\
\sin (\theta) & \cos(\theta) & 0\\
0 & 0 & 1
\end{bmatrix}    \begin{bmatrix}
        x \\ y \\ 1
    \end{bmatrix}.
    \label{equation:coordinate_transfer}
\end{equation}

Finally, we map the relative coordinates to GPS coordinates via
\begin{align}
    la^{object} &= la + \frac{y_r}{r} \frac{180}{\pi}, \\
    lo^{object} &= lo + \frac{x_r}{r} \frac{180}{\pi} \frac{1}{\cos (lat \ \pi / 180)}.
\end{align}

We refer the reader to \cite{kiefer2023memory} for a more comprehensive derivation of the 2D $\longleftrightarrow$ 3D projection. Concretely, we would like to note that the projection may especially be critical for a distant object in the USV scenario as here, we encounter a very acute viewing angle. Small errors in pixel space result in large distance errors in world space. It is an open problem of how to correctly project distant objects in world space. For our consideration, we are mostly concerned with obtaining correct heading estimations for either close detections that may ultimately pose an immediate threat. For distant objects, we mostly care about stable heading predictions over time.

\section{Data Collection and Labeling}
\label{sec:datacollection_labeling}

Because of the lack of available datasets for yaw estimation, we capture and annotate our own. For the UAV scenario, we leverage the already existing SeaDronesSee-MOT \cite{varga2021seadronessee} dataset, which comes with bounding boxes and instance ids for boats. Furthermore, we annotate the 6D pose of boats from various sample scenes by adapting the annotation tool provided in \cite{xiang2014beyond}. Figure \ref{fig:3dlabel_tool} shows an example scene where a boat is labelled from a viewpoint of a UAV. We leverage the provided metadata from the UAV to automatically infer coarse pitch and roll angles relative to the camera. Herein, we assume the world pitch and roll angle of boats to be zero, such that we only need to annotate the heading direction. For that, we manually provide a coarse heading and, upon selecting anchor points from the CAD in the corresponding real objects, we optimize for the precise 6D pose using their optimization procedure \cite{xiang2014beyond}. For annotation efficiency, we only annotate every 10th frame and interpolate the pose in between. 

\begin{figure}
\vspace{-3mm}
    \centering
    \includegraphics[trim=0 0 0 0,clip,width=0.98\textwidth]{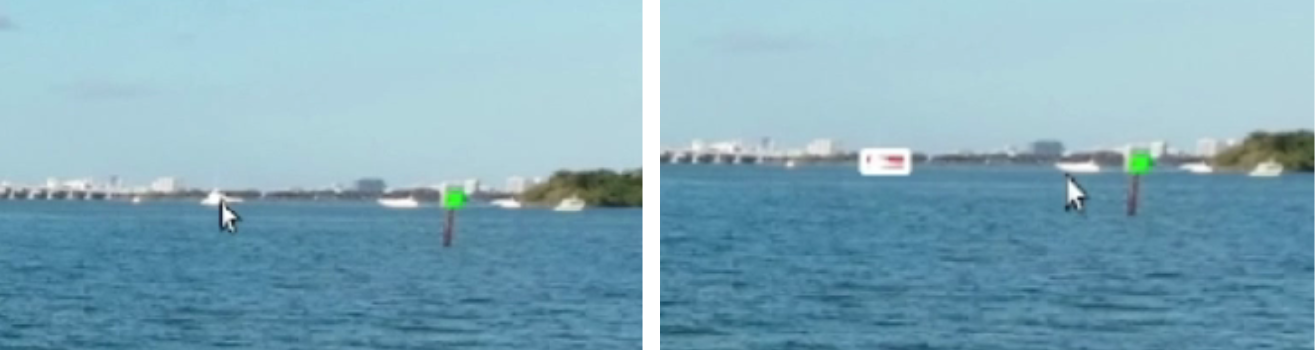}
    
    \caption{Faster bounding box annotations by means of "Segment anything" \cite{kirillov2023segany}. We leverage this method to accelerate 2D bounding box labeling. A user just needs to click on the object, the corresponding bounding box will be set and saved automatically.}
    \label{fig:3dlabel_tool_ground}
    
\end{figure}

For the USV scenario, we capture our own data from the viewpoint of a fixed camera installed on a small motorboat. We use the ZED2 camera\footnote{\url{https://www.stereolabs.com/zed-2/}} with integrated IMU to infer the orientation at which we look at the scene. As before, we may also infer a coarse estimation of the roll and pitch angle for subsequent finer annotating via pose optimization. Before that, we annotate the scenes with bounding boxes using our tool, which we built on top of SAM (Segment Anything Model \cite{kirillov2023segany}, see Fig. \ref{fig:3dlabel_tool_ground}). We leveraged their largest ViT-H (636M parameters) model and built a user interface and labeling logic around it, such that objects can be assigned their bounding boxes by just clicking on them. Analogous to before, we annotate every 10th frame and interpolate in between.  Table \ref{table:annotation_tool_speedup} shows a timing comparison between conventional labeling tools and our method. Every method was required to yield precise bounding boxes as rated by human experts.  
We repeated this experiment with five experts knowledgeable in the field of object detection. Each experiment lasted for half an hour. Our method clearly outperforms the others by 8.7 FPM. We hypothesize that fatigue symptoms occur later because annotating with a single click already covers the entire object. In contrast, when setting bounding boxes, precise outlining of the object is required, which becomes more exhausting over time. While this effort can be reduced by tracking, there are often errors in tracking objects that are far away, requiring the annotator to stay alert and relabel bounding boxes.

We want to note that this method can fail in scenarios of low contrast or very distant objects. In this case, one has to resort to standard bounding box detection. Moreover, it requires a GPU to process ViT-H (in our case an RTX 2080Ti). Furthermore, a more exhaustive study on the benefits of segmentation-based labeling needs to be done to obtain a more comprehensive overview. In particular, a more comprehensive experiment considering object number, size, shape and movement needs to be done. We release both (adapted) annotation tools for further studies.

\begin{table}	
	\begin{center}		
		\begin{tabular}{lr}
                \toprule
			Annotation method  & Labeling Speed (FPM)  \\
			\midrule
                DarkLabel \cite{darklabel} &  3.8 \\
                DarkLabel + Interpolation &  19.6 \\ 
                DarkLabel + Tracking &  20.2 \\                
                SAM-based & 5.5 \\
                \bf SAM-based + Interpolation & \bf 28.9 \\
                \bottomrule
		\end{tabular}
	\end{center}
	\caption{Annotation speed given in frames per minute using our annotation tool based on SAM and ViT-H. }
	\label{table:annotation_tool_speedup}
\end{table}

\section{Method: Aggregating Probability Distributions over Time}


Our approach is based on HyperPosePDF \cite{murphy2021implicit}. For an input image $x \in \mathcal{X}$, it aims to obtain a conditional probability distribution $\fun{p(\cdot | x)}{\SO(3)}{\R^+}$, representing the distribution of the inherited pose of an object in the image $x$. For that, we train a vision backbone network, e.g. ResNet to predict the networks of a second network. While the vision network acts as a hypernetwork, the architecture of the second network is inspired by an implicit neural representation. The implicit neural representation acts on the rotation manifold and outputs for each pose, the corresponding probability of it being the underlying rotation of the object present in the image. Hence, it acts as an approximation of the probability distribution $p(R|x)$ by marginalizing over $\SO(3)$. During training, we maximize $p(R|x)$ by providing pairs of inputs $x$ and corresponding ground truth $R$. To make a single pose prediction, we solve $\argmax_{R \in \SO(3)} f(x, R)$ with gradient ascent, projecting the values back into the manifold after each step. To predict a full probability distribution, we evaluate $p(R_i|x)$ over the \SO(3) equivolumetric partition ${R_i}$. This model can estimate complex distributions on the manifold without prior knowledge of each object's symmetries, and appropriate patterns expressing symmetries and uncertainty emerge naturally in the model's outputs. This is indeed, the most general way to conduct pose estimation. Specifically, in our scenario where we want to predict the pose to predict the trajectory it is possible to include uncertainty information of the pose to improve the performance.

The posterior distribution
\begin{align}
    P(R_{k+1} | Z_k),
\end{align}
based on the observations $Z_k$ for time steps $\{1,\dots,k\}$ can be approximated by 
\begin{align}
    P(R_{k+1} | Z_k) \approx P(R_k|Z_k) + \Delta_{\text{pose}},
\end{align}
where $\Delta_{\text{pose}}$ is defined as a weighted running average
\begin{align}
    \Delta_{\text{pose}} = \frac{1}{k} \sum_{l=0}^{k-1} \omega_l\Big(P(R_{l+1}|Z_{l+1}) - P(R_l| Z_l)\Big).
\end{align}
  For $l < k+1$ the probabilities $P(R_l| Z_l)$ are known and approximated by the HyperPosePDF network. Therefore, the calculation of the pose at a future time point is deterministic. The weights $\omega_l$ fulfill $\sum_l \omega_l = 1$. To reduce the effect of earlier pose transitions, which have a lesser effect on the current pose movement it is plausible to simply set the initial weights as $0$ and average the remaining over a smaller time interval chosen such that $0<t<k$
  \begin{align}
    \omega_l = \begin{cases}0 \;\;\;\;\;\;\;\; &\text{ for } l < t, \\
    \frac{1}{k-t} \;\;\;\;\; &\text{ for } l \ge t.
    \end{cases}
  \end{align}
This especially comes in helpful, when we try to predict the movement of a boat that is in the middle of a turn manoeuvre and the respective trajectory resembles a curve. Furthermore, this allows us to detect false pose predictions in the case that the pose prediction in the next time step differs to much from the previous path. E.g., in the case of nearly symmetric boats, we experienced the appearance of $180^\circ$ miss-predictions, which now can be easily excluded.




\section{Experiments}



First, we conduct experiments on the single-image Pascal3D+ set to illustrate the performance and expressiveness of HyperPosePDF. Similar to \cite{Hofer_2023_WACV}, we choose a pretrained ResNet-101 backbone for our vision module. Then, we train the model to predict the weights of a one-layer network with a width of 256. Using the Adam optimizer, we evaluate our model after 150k iterations using a batch size of 16. A learning rate of $10^{-5}$ is used for the first 1000 iterations, and then a cosine decay
is applied. We choose a time horizon window of $k=3$ for our experiments.

We report the performance of the category \emph{boat} using the two commonly used metrics \emph{accuracy at 30$^\circ$ (Acc)} and \emph{median error in degrees (ME)}. Table \ref{table:pose_estimation} shows that this method is on par with SOTA methods (ImplicitPDF \cite{murphy2021implicit}).

\begin{figure*}
    \centering
    \includegraphics[width=0.24\textwidth,trim={0 0 470 0},clip]{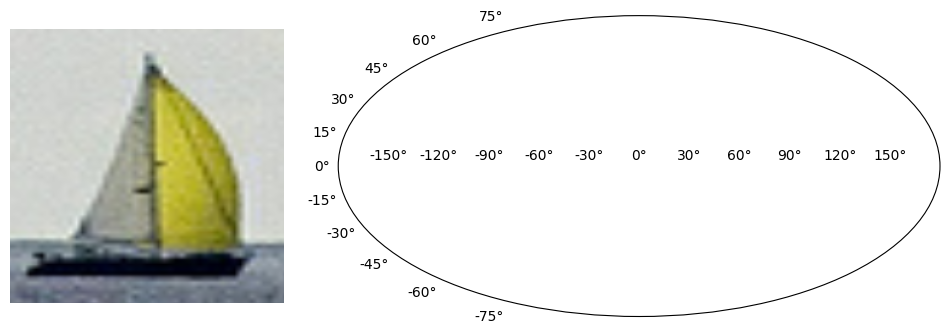}
    \includegraphics[width=0.24\textwidth,trim={0 0 470 0},clip]{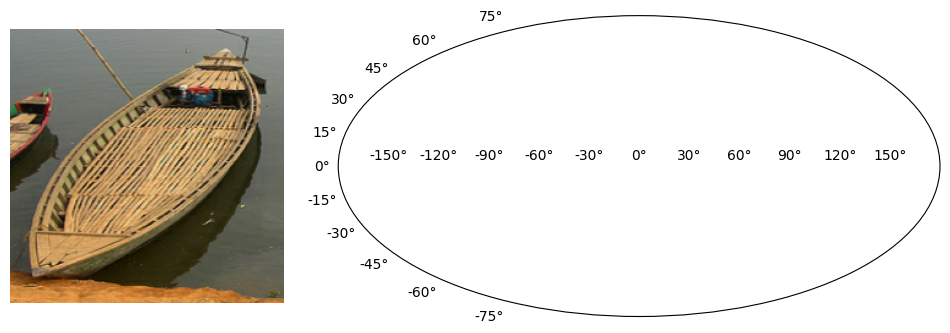}
    \includegraphics[width=0.24\textwidth,trim={0 0 470 0},clip]{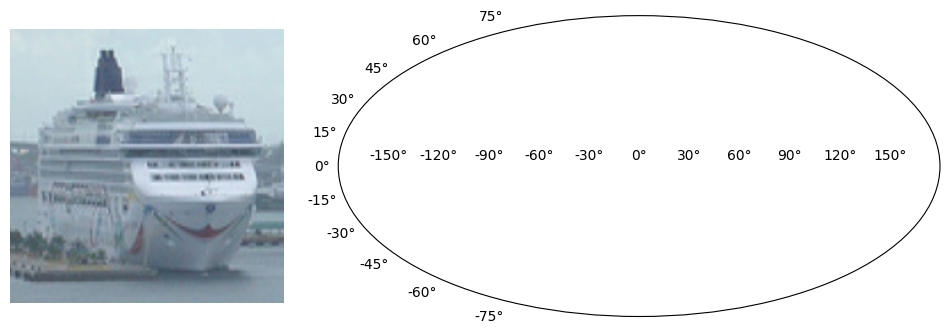}
    \includegraphics[width=0.24\textwidth,trim={0 0 470 0},clip]{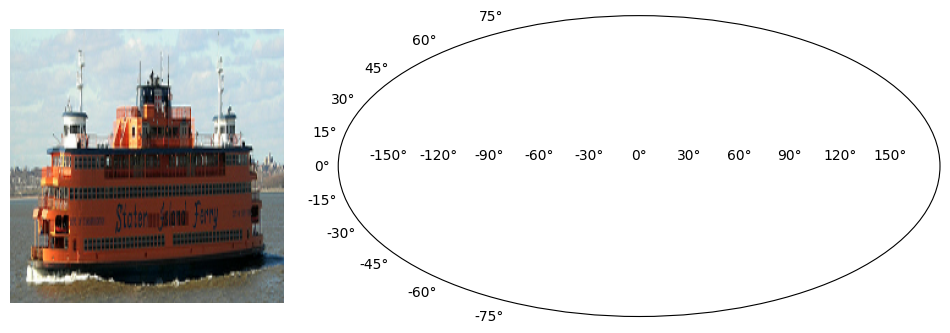}\\
    
    \includegraphics[width=0.23\textwidth,trim={35 0 10 200},clip]{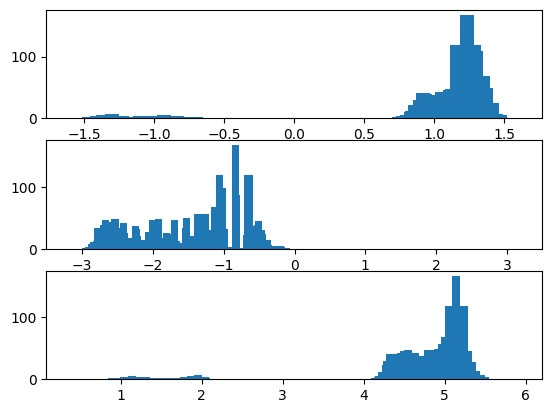}
    \includegraphics[width=0.23\textwidth,trim={40 0 10 200},clip]{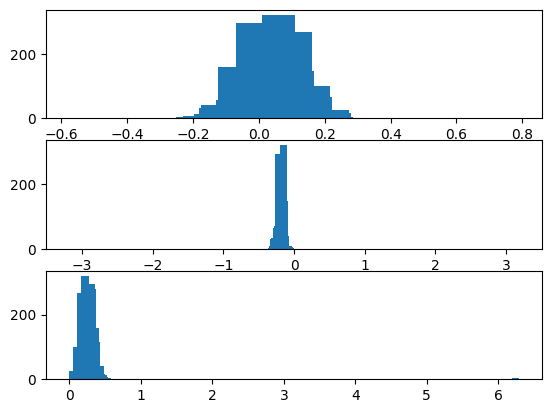}
    \includegraphics[width=0.23\textwidth,trim={40 0 10 200},clip]{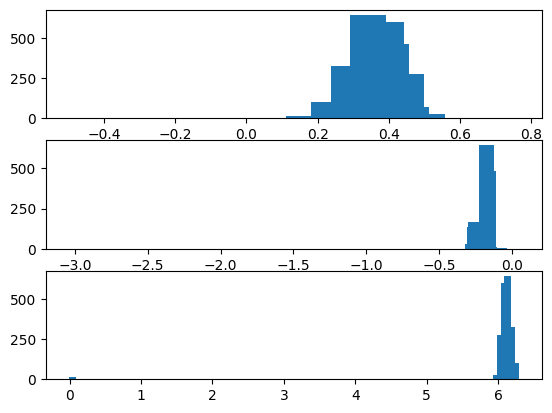}
    \includegraphics[width=0.23\textwidth,trim={35 0 10 200},clip]{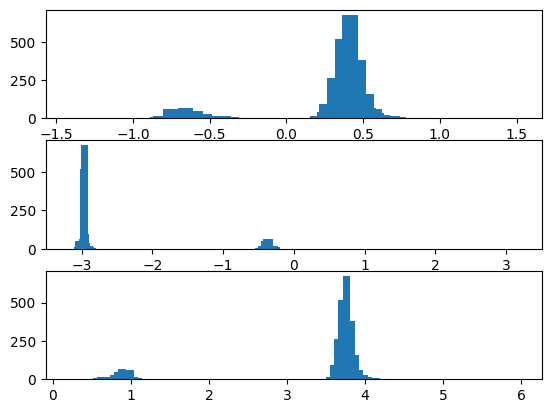}
    \caption{Sample boat heading probability distribution predictions (given in radian). Ground truth values are 4.8, 0.4, 6.0, 4.0. Almost all the captured distributions are uni-modal and the best single-point estimator would yield a fairly close prediction of the orientation. The first image already provides a glance at the benefits of capturing the uncertainty. It is not clear whether the sailboat is sailing at an angle of 270$^\circ$ or slightly less.}
    \label{fig:distribution_pascal_imgs}
\end{figure*}

Naively applying HyperPosePDF on video data yields single-point predictions (i.e. orientations) at each time step. However, uncertainty due to unseen data yields far-off predictions, often resulting in wrong symmetric orientations. For example, compare to Figure \ref{fig:pose_estimation}.

We evaluate on SeaDronesSee3D and BOArienT, where we manually annotated the orientations. Table \ref{table:pose_estimation} shows that our method yields higher accuracy at a maximum of $30^\circ$ error tolerance as well as lower median angle error. Figure \ref{fig:pose_estimation} shows an example sequence of SeaDronesSee3D where the single-image predictor miss-predicts the orientation by 180$^\circ$ due to the slight symmetric shape of the boat. 

\begin{table}	
	\begin{center}		
		\begin{tabular}{lrrr}
                \toprule
			Method &  Dataset & Acc$\uparrow$ & ME$\downarrow$  \\
			\midrule
                ImplicitPDF  & PASCAL3D+ & 56.0 & \bf 23.4 \\
                HyperPosePDF  & PASCAL3D+ & \bf 56.2 & 22.8 \\
            \midrule
                HyperPosePDF  & SeaDronesSee3D & 65.6 & 20.1 \\
                \bf +Run. Mean & SeaDronesSee3D & \bf 71.9 &  \bf 16.7\\
            \midrule
                HyperPosePDF & BOArienT & 42.5 & 41.8 \\
                \bf +Run. Mean & BOArienT & \bf 50.2 &  \bf 18.3\\
                
                \bottomrule
		\end{tabular}
	\end{center}
	\caption{Yaw estimation results on PASCAL3D+, SeaDronesSee3D and BOArienT. Note that PASCAL3D+ only features still images.}
 \vspace{-5mm}
	\label{table:pose_estimation}
 
\end{table}



To test our approach in a complete pipeline, we employ a state-of-the-art multi-object tracker and apply the yaw estimator on the predicted bounding boxes. For the UAV scenario, we train on SeaDronesSee-MOT, and for the boat scenario, we take a pre-trained tracker on COCO. We report the performance of the trackers on SeaDronesSee3D and BOArienT in Table \ref{table:accuracy_tracking}.

\begin{table}	
	\begin{center}		
		\begin{tabular}{llrrrr}
                \toprule
			& Model  &  HOTA$\uparrow$ & MOTA$\uparrow$ &  IDs$\downarrow$ & Frag$\downarrow$  \\
			\midrule
              {\multirow{3}{*}{\rotatebox[origin=c]{90}{SDS3D}}}  & ByteTracker & 79.9 & 89.8 & 23 & 678   \\
               & DeepSORT & 80.8 & 91.0 & 20 & 642 \\
                & \bf +MM & \bf 82.6 & \bf 91.9 & \bf 19 & \bf 635 \\
                \midrule
              {\multirow{2}{*}{\rotatebox[origin=c]{90}{BT}}}  & Tracktor  & 65.6  & 67.0 & 69 & 876   \\
               & DeepSORT  & 66.2 & 80.0 & 51 & 801 \\
                \bottomrule
		\end{tabular}
	\end{center}
	\caption{Multi-Object Tracking accuracy on SeaDronesSee3D (SDS3D) and BOArienT (BT). For SDS3D, we used the methods from the workshop competition \cite{kiefer20231st}. Additionally, we built on top DeepSORT a memory map (MM) \cite{kiefer2023memory} to become more robust towards id switches and fragmentations. For BT, we used off-the-shelf Tracktor \& DeepSORT.}
	\label{table:accuracy_tracking}
 \vspace{-5mm}
\end{table}

Now, we apply the yaw estimator on top of the predicted bounding boxes with associated ids. Whenever a new tracklet is starting, we initialize a new probability distribution running mean. We only measure the orientation estimation performance on objects that have successfully been detected.

\begin{table}	
	\begin{center}		
		\begin{tabular}{llrrr}
                \toprule
			& Method+Tracking  & Acc$\uparrow$ & ME$\downarrow$  \\
			\midrule
               {\multirow{3}{*}{\rotatebox[origin=c]{90}{SDS3D}}} & Trajectory-based &   23.1 & 123.3\\
                & HyperPosePDF &   63.2 & 22.1 \\
                & +Mode Running Mean   & 64.3 & 21.6 \\
                &\bf +Distribution Running Mean   & \bf 70.3 &  \bf 17.3\\
            \midrule
             {\multirow{3}{*}{\rotatebox[origin=c]{90}{BOArienT}}} & Trajectory-based   & 20.2 & 72.6\\
               & HyperPosePDF   & 39.6 & 43.9 \\
               & +Mode Running Mean   & 40.1 & 42.0 \\
               & \bf +Distribution Running Mean   & \bf 49.8 &  \bf 19.5\\
                
                \bottomrule
		\end{tabular}
	\end{center}
	\caption{Yaw estimation results on SeaDronesSee3D and BOArienT.}
	\label{table:pose_estimation_based_on_tracking}
 \vspace{-5mm}
\end{table}

Table \ref{table:pose_estimation_based_on_tracking} shows that our method still outperforms the single-image approach since the multi-object tracker is quite robust already (only a few ID switches degrade our method to effectively become a single-image method at these time points). Remarkably, we can even improve the point prediction over the naive mode running mean method, which simply applies a running mean on the modes of the distributions. We note, that this is on top of the higher expressiveness coming from our probability distributions: we may incorporate the uncertainty of heading estimations in downstream tasks, such as trajectory prediction, collision avoidance or for visualization purposes in augmented reality applications.

Finally, we compare our heading estimation approach with a naive trajectory-based approach. For every detection in every frame, we map its center box location to 3D via a perspective projection camera model \cite{kiefer2023memory} and capture the trajectory in world coordinates. We predict the next trajectory point by a constant velocity model coming from the previous three time steps. We take the resulting heading to be the final prediction of this baseline. If an object is lost, we need to re-initialize the heading which is a critical shortcoming of this approach. Furthermore, Table \ref{table:pose_estimation_based_on_tracking} shows that the trajectory-based method fails on both scenarios due to stationary boats and a challenging and noisy 2D $\rightarrow$ 3D projection.

\begin{figure}
    \centering
    \includegraphics[width=0.98\textwidth]{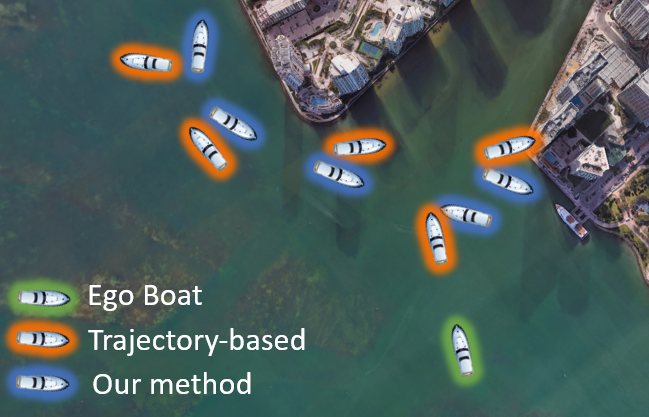}
    \caption{Sample synthetic rendering of the scene from Figure \ref{fig:example_orientation}. Detected boats and their heading are put into Google Earth. The big yacht on the right was already contained in the Google Earth image. We add a slight offset to the predicted locations (which are the same for the two methods) for visualization purposes.\vspace{-3mm}}
    \label{fig:boats_top_down}
    \vspace{-2mm}
\end{figure}

Figure \ref{fig:boats_top_down} shows the predicted positions and headings in BOArienT coming from our method and from this baseline via 2D $\rightarrow$ 3D projection. 
Because some boats are stationary, the heading information for the baseline is incorrect. Furthermore, the heading information from slowly driving boats is very noisy as the underlying 2D $\longleftrightarrow$ 3D projection is error-prone. Single-image predictions are better, but the smallness of the objects makes these predictions also very noisy.


\section{Conclusion and Discussion}

In this paper, we addressed the novel problem of predicting the yaw of boats from the viewpoint of unmanned aerial vehicles (UAVs) and unmanned surface vehicles (USVs) or boats. We proposed a method based on HyperPosePDF, which models the uncertainty of predictions and yields robust orientation predictions across time in video-based scenarios. To demonstrate the utility of our method, we created two new datasets, SeaDronesSee-3D and BOArienT, manually annotated with bounding boxes and pose information, and made them publicly available. Our experimental evaluation showed that our method significantly improves performance compared to naive single-point predictions. Our proposed method has potential applications in marine robotics, including 3D scene rendering, trajectory prediction, and navigation.

\bibliographystyle{IEEEtran}
\bibliography{IEEEabrv,IEEEexample}

\begin{thebibliography}{10}
\providecommand{\url}[1]{#1}
\csname url@rmstyle\endcsname
\providecommand{\newblock}{\relax}
\providecommand{\bibinfo}[2]{#2}
\providecommand\BIBentrySTDinterwordspacing{\spaceskip=0pt\relax}
\providecommand\BIBentryALTinterwordstretchfactor{4}
\providecommand\BIBentryALTinterwordspacing{\spaceskip=\fontdimen2\font plus
\BIBentryALTinterwordstretchfactor\fontdimen3\font minus
  \fontdimen4\font\relax}
\providecommand\BIBforeignlanguage[2]{{%
\expandafter\ifx\csname l@#1\endcsname\relax
\typeout{** WARNING: IEEEtran.bst: No hyphenation pattern has been}%
\typeout{** loaded for the language `#1'. Using the pattern for}%
\typeout{** the default language instead.}%
\else
\language=\csname l@#1\endcsname
\fi
#2}}

\bibitem{arnold2019survey}
E.~Arnold, O.~Y. Al-Jarrah, M.~Dianati, S.~Fallah, D.~Oxtoby, and
  A.~Mouzakitis, ``A survey on 3d object detection methods for autonomous
  driving applications,'' \emph{IEEE Transactions on Intelligent Transportation
  Systems}, vol.~20, no.~10, pp. 3782--3795, 2019.

\bibitem{Hofer_2023_WACV}
T.~H\"ofer, B.~Kiefer, M.~Messmer, and A.~Zell, ``Hyperposepdf - hypernetworks
  predicting the probability distribution on so(3),'' in \emph{Proceedings of
  the IEEE/CVF Winter Conference on Applications of Computer Vision (WACV)},
  January 2023, pp. 2369--2379.

\bibitem{kirillov2023segany}
A.~Kirillov, E.~Mintun, N.~Ravi, H.~Mao, C.~Rolland, L.~Gustafson, T.~Xiao,
  S.~Whitehead, A.~C. Berg, W.-Y. Lo, P.~Doll{\'a}r, and R.~Girshick, ``Segment
  anything,'' \emph{arXiv:2304.02643}, 2023.

\bibitem{hofer2021object}
T.~H{\"o}fer, F.~Shamsafar, N.~Benbarka, and A.~Zell, ``Object detection and
  autoencoder-based 6d pose estimation for highly cluttered bin picking,'' in
  \emph{2021 IEEE International Conference on Image Processing (ICIP)}.\hskip
  1em plus 0.5em minus 0.4em\relax IEEE, 2021, pp. 704--708.

\bibitem{labbe2020cosypose}
Y.~Labb{\'e}, J.~Carpentier, M.~Aubry, and J.~Sivic, ``Cosypose: Consistent
  multi-view multi-object 6d pose estimation,'' in \emph{Computer Vision--ECCV
  2020: 16th European Conference, Glasgow, UK, August 23--28, 2020,
  Proceedings, Part XVII 16}.\hskip 1em plus 0.5em minus 0.4em\relax Springer,
  2020, pp. 574--591.

\bibitem{sundermeyer2020augmented}
M.~Sundermeyer, Z.-C. Marton, M.~Durner, and R.~Triebel, ``Augmented
  autoencoders: Implicit 3d orientation learning for 6d object detection,''
  \emph{International Journal of Computer Vision}, vol. 128, pp. 714--729,
  2020.

\bibitem{murphy2021implicit}
K.~Murphy, C.~Esteves, V.~Jampani, S.~Ramalingam, and A.~Makadia,
  ``Implicit-pdf: Non-parametric representation of probability distributions on
  the rotation manifold,'' \emph{arXiv arXiv:2106.05965}, 2021.

\bibitem{varga2021seadronessee}
L.~A. Varga, B.~Kiefer, M.~Messmer, and A.~Zell, ``Seadronessee: A maritime
  benchmark for detecting humans in open water,'' \emph{arXiv preprint
  arXiv:2105.01922}, 2021.

\bibitem{kiefer2023fast}
B.~Kiefer and A.~Zell, ``Fast region of interest proposals on maritime
  {UAV}s,'' \emph{arXiv preprint arXiv:2301.11650}, 2023.

\bibitem{MODSBenchmark2022}
B.~Bovcon, J.~Muhovič, D.~Vranac, D.~Mozetič, J.~Perš, and M.~Kristan,
  ``{MODS--A USV-Oriented Object Detection and Obstacle Segmentation
  Benchmark},'' \emph{IEEE Transactions on Intelligent Transportation Systems},
  pp. 1--16, 2021.

\bibitem{kiefer20231st}
B.~Kiefer, M.~Kristan, J.~Per{\v{s}}, L.~{\v{Z}}ust, F.~Poiesi, F.~Andrade,
  A.~Bernardino, M.~Dawkins, J.~Raitoharju, Y.~Quan, \emph{et~al.}, ``1st
  workshop on maritime computer vision (macvi) 2023: Challenge results,'' in
  \emph{Proceedings of the IEEE/CVF Winter Conference on Applications of
  Computer Vision}, 2023, pp. 265--302.

\bibitem{sullivan2006predictive}
B.~Sullivan, C.~Ware, and M.~Plumlee, ``Predictive displays for survey
  vessels,'' in \emph{Proceedings of the Human Factors and Ergonomics Society
  Annual Meeting}, vol.~50, no.~22.\hskip 1em plus 0.5em minus 0.4em\relax Sage
  Publications Sage CA: Los Angeles, CA, 2006, pp. 2424--2428.

\bibitem{browning1991mathematical}
A.~W. Browning, ``A mathematical model to simulate small boat behaviour,''
  \emph{Simulation}, vol.~56, no.~5, pp. 329--336, 1991.

\bibitem{du2018unmanned}
D.~Du, Y.~Qi, H.~Yu, Y.~Yang, K.~Duan, G.~Li, W.~Zhang, Q.~Huang, and Q.~Tian,
  ``The unmanned aerial vehicle benchmark: Object detection and tracking,'' in
  \emph{Proceedings of the European Conference on Computer Vision (ECCV)},
  2018, pp. 370--386.

\bibitem{s23073691}
\BIBentryALTinterwordspacing
P.~Ruiz-Ponce, D.~Ortiz-Perez, J.~Garcia-Rodriguez, and B.~Kiefer, ``Poseidon:
  A data augmentation tool for small object detection datasets in maritime
  environments,'' \emph{Sensors}, vol.~23, no.~7, 2023. [Online]. Available:
  \url{https://www.mdpi.com/1424-8220/23/7/3691}
\BIBentrySTDinterwordspacing

\bibitem{kiefer2021leveraging}
B.~Kiefer, D.~Ott, and A.~Zell, ``Leveraging synthetic data in object detection
  on unmanned aerial vehicles,'' \emph{arXiv preprint arXiv:2112.12252}, 2021.

\bibitem{messmer2021gaining}
M.~Messmer, B.~Kiefer, and A.~Zell, ``Gaining scale invariance in uav bird's
  eye view object detection by adaptive resizing,'' \emph{arXiv preprint
  arXiv:2101.12694}, 2021.

\bibitem{kiefer2022leveraging}
B.~Kiefer, D.~Ott, and A.~Zell, ``Leveraging synthetic data in object detection
  on unmanned aerial vehicles,'' in \emph{2022 26th International Conference on
  Pattern Recognition (ICPR)}.\hskip 1em plus 0.5em minus 0.4em\relax IEEE,
  2022, pp. 3564--3571.

\bibitem{yawpitchroll}
\BIBentryALTinterwordspacing
Jmvolc. Rotations around axes. [Online]. Available:
  \url{https://en.wikipedia.org/wiki/Ship_motions#/media/File:Rotations.png}
\BIBentrySTDinterwordspacing

\bibitem{kiefer2023memory}
B.~Kiefer, Y.~Quan, and A.~Zell, ``Memory maps for video object detection and
  tracking on uavs,'' \emph{arXiv preprint arXiv:2303.03508}, 2023.

\bibitem{xiang2014beyond}
Y.~Xiang, R.~Mottaghi, and S.~Savarese, ``Beyond pascal: A benchmark for 3d
  object detection in the wild,'' in \emph{IEEE winter conference on
  applications of computer vision}.\hskip 1em plus 0.5em minus 0.4em\relax
  IEEE, 2014, pp. 75--82.

\bibitem{darklabel}
U.~darkpgmr, ``{DarkLabel Annotation Tool, Github},''
  \url{https://github.com/darkpgmr/DarkLabel}, accessed: 2022-07-05.

\end{thebibliography}

\end{document}